\documentclass[10pt,twocolumn,letterpaper]{article}

\usepackage{ijcb}
\usepackage{times}
\usepackage{epsfig}
\usepackage{graphicx}
\usepackage{amsmath}
\usepackage{amssymb}
\usepackage{physics}
\usepackage{subcaption}

\ijcbfinalcopy % *** Uncomment this line for the final submission

\ifijcbfinal\fi
\begin{document}

%%%%%%%%% TITLE
\title{Are Gabor Kernels Optimal for Iris Recognition?}

% Authors at the same institution
%\author{First Author \hspace{2cm} Second Author \\
% Authors at different institutions
\author{Aidan Boyd, Adam Czajka, Kevin Bowyer \\
University of Notre Dame\\
{\tt\small \{aboyd3, aczajka, kwb\}@nd.edu}
% \and
% Second Author \\
}

\maketitle
\thispagestyle{empty}

%%%%%%%%% ABSTRACT
\begin{abstract}

Gabor kernels are widely accepted as dominant filters for iris recognition. In this work we investigate, given the current interest in neural networks, if Gabor kernels are the only family of functions performing best in iris recognition, or if better filters can be learned directly from iris data. We use (on purpose) a single-layer convolutional neural network as it mimics an iris code-based algorithm. We learn two sets of data-driven kernels; one starting from randomly initialized weights and the other from open-source set of Gabor kernels. Through experimentation, we show that the network does not converge on Gabor kernels, instead converging on a mix of edge detectors, blob detectors and simple waves. In our experiments carried out with three subject-disjoint datasets we found that the performance of these learned kernels is comparable to the open-source Gabor kernels. These lead us to two conclusions: (a) a family of functions offering optimal performance in iris recognition is wider than Gabor kernels, and (b) we probably hit the maximum performance for an iris coding algorithm that uses a single convolutional layer, yet with multiple filters. Released with this work is a framework to learn data-driven kernels that can be easily transplanted into open-source iris recognition software (for instance, OSIRIS -- Open Source IRIS).
\end{abstract}

\begin{figure}[t]
    \centering
    \includegraphics[width=1\columnwidth]{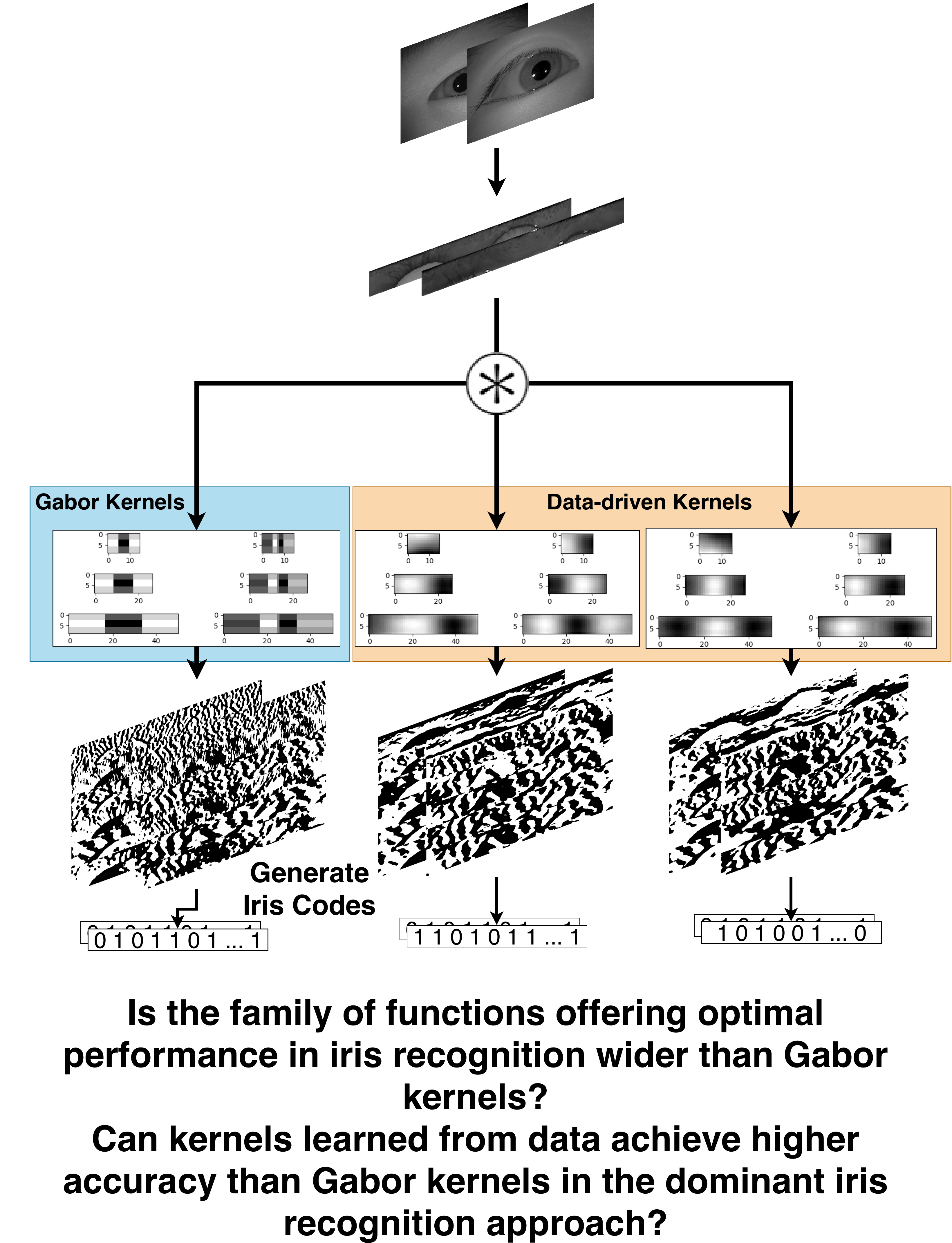}
    \caption{High-level overview of the proposed methodology. Iris images are segmented and normalized and then convolved with three sets of kernels independently. The first set are hand-crafted Gabor kernels and the other two sets are data-driven kernels learned with a shallow network. The resulting encodings are then sampled to produce an iris code for the image. These iris codes are used to perform matching.}
    \label{teaser}
\end{figure}
%%%%%%%%% BODY TEXT
\section{Introduction}

Gabor kernels are widely recognized as an effective tool for iris feature extraction, as proposed by Daugman \cite{Daugman_PAMI_1993}, and still are a dominant approach for calculating iris codes in iris recognition systems. Daugman's method is straightforward in principle: convolve a normalized iris image with a set of kernels (only real parts of Gabor kernels are used in commercial systems \cite{daugman2015information}), binarize the result to form a code, and use only ``strong'' (not ``fragile'') bits corresponding to non-occluded parts of the iris in calculating the distance between two iris samples. An important question is how to find optimal kernels that offer highly individual iris codes. Should these kernels be selected only from a family of Gabor kernels, as in a dominant approach, or can they be selected from a broader set of functions? And if we start from a broader set of kernels and run a hyperparameter search, will we eventually (and independently of the database) converge to Gabor kernels? Since they have solid theoretical foundations in computer vision (e.g., they serve as uncertainty-minimising elementary functions, and have been found to model activation profiles of visual cortex neurons \cite{jones1987evaluation}), observing that such a hyperparameter search always converges to Gabor kernels would simply justify our choice for using these kernels in iris recognition. If, however, optimal filtering kernels turn out to be different than Gabor, what do they look like? And how to find them effectively given a database of iris images? Do they perform better than Gabor, or worse? This paper answers these questions.

We propose to use a single-layer convolutional neural network that replicates a Daugman's approach to iris recognition. This network provides a framework to learn data-driven kernels for iris recognition that can be directly implanted into open-source tools such as OSIRIS \cite{othman2016osiris}. The input to the network is a normalized iris image and the output is a feature vector of size $1536$ which mimics an iris code. Figure \ref{teaser} shows the high-level overview of the pipeline. The iris occlusion mask is incorporated directly into training, so the network does not use non-iris areas in searching for optimal filters. We also propose a triplet loss function which incorporates normalized occlusion masks into the loss and also uses a domain-specific distance metric and a soft margin \cite{hermans2017defense}. We considered starting the training procedure from randomly initialized weights, and from weights initialized with open-source Gabor kernels.  

The main and interesting {\bf findings} presented in this paper are:

\begin{itemize}
    \item the network training procedure converges to kernels deviant of Gabor kernels, instead converging on a combination edge detectors and blob detectors, 
    \item the above happens independently of the weight initialization procedure (initialization with Gabor kernels or random numbers),
    \item these learned kernels are also shown to offer similar performance than the only known open-source Gabor kernels included with the OSIRIS tool.
\end{itemize}

The data used to train the kernels was a large set of in-house data of more than 340,000 iris images collected from 3,000 distinct eyes. For testing result reporting purposes we used three independent datasets of varying difficulty:
\begin{itemize}
    \item {\it CASIA-Iris-Thousand V4},
    \item {\it WVU Non-ideal Iris Database – Release 1},
    \item {\it Live} partition from the entire {\it LivDet 2017 Liveness Detection-Iris -- Warsaw Subset}.
\end{itemize}

% For the purpose of generalization, the datasets used to test the data-driven kernels in this work are 
% Reported in this paper is the accuracy of the data-driven kernels in comparison with Gabor kernels using the scores from the OSIRIS tool.

{\bf Practical contributions} of this paper include:

\begin{itemize}
    \item kernels that are ready to be used in any implementation of Daugman's algorithm ({\it e.g.}, OSIRIS), that were learned using a large database of 340,000 iris images collected from 3,000 classes,
    \item source codes of the framework allowing to train filters for own database of iris images along with all visualization scripts \cite{aboyd-github}. % {https://github.com/BoydAidan/ExploringGaborFiltersIJCB2020}

\end{itemize}

% -- We think Gabor best, but what does a neural network think?
% \subsection{Our Contributions}
% -- New set of kernels
% -- Show that Gabor may not be optimal
% -- Adam: you had ideas for this right?
% -- First paper to incorporate masks directly into singular network?

% The main contributions of this paper include: a framework to generate data-driven kernels for iris recognition, a new set of kernels learned using a large iris database that show improvements over open-source Gabor kernels and a methodology to include occlusion masks in the training of neural networks

\section{Related Work}

In a work by Zhao \etal \cite{zhao2017towards}, the use of triplet loss for the purpose of iris recognition is explored. This paper outlines the increase in performance and generalization that can be attained by applying a modified triplet loss that includes occlusion masks to the task of iris recognition. Direct comparisons are made to the performance to the 2D Gabor kernels included with the OSIRIS tool. {\bf Our paper differs from this} in that instead of creating a multi-layer network to learn the kernels, we create a one layer network to learn the \textit{kernels} that can be used outside of the model and accurately mimic the Daugman approach to iris recognition. Their work was later extended \cite{wang2019towards} to show performance increases using a dilated neural network. The purpose of their work was to increase performance of an iris recognition system, whereas we wish to apply iris data to a one layer convolutional network and examine the kernels that are learned to determine whether Gabor kernels are optimal. Additionally, our work applies the occlusion masks directly to the loss function, rather than passing through a network as in the above works.% We also investigate whether weight initialization has an effect on performance and convergence whereas in their works, the deep networks were initialized randomly. 

The use of batch mining in our work is described in a work by Hermans \etal \cite{hermans2017defense}. In their work, extensive experimentation is done on different triplet selection methodologies and they show that the use of hard mining within batches (referred to as \textit{Batch Hard}) was optimal for their application to person re-identification. We adapt this approach to our task and perform Batch Hard selection of triplets. Also described in this paper is the use of a soft-margin for triplet loss calculation. They describe that instead of using an $\alpha$ parameter to determine separation between classes within triplets, a log based approach provided a smoother decay without a cutoff as with the original approach. In our work we use this soft-margin to calculate the triplet loss.

In a work by Ahmad and Fuller \cite{ahmad2019thirdeye}, both the Batch Hard mining and soft-margin approach is explored for the purpose of iris recognition without normalization. In this work, this approach to triplet loss is validated by showing that a triplet based network can achieve better performance than end to end deep networks such as DeepIrisNet \cite{gangwar2016deepirisnet}. {\bf Our work is different from this} in that we use normalized images as the input to the network and include the occlusion mask in the loss function. In their work, a five convolutional layer structure is proposed with a pooling layer. No kernel analysis is performed. In our work we analyse the kernels learned by our single convolutional layer.

% Hermans \etal propose the use of a soft-margin within the triplet loss function to 
% The application of batch mining for triplet loss applications, as described by Hermans \etal 

\section{Methodology}

\subsection{Databases}

\begin{figure}[t]
    \centering
    \includegraphics[width=1\columnwidth]{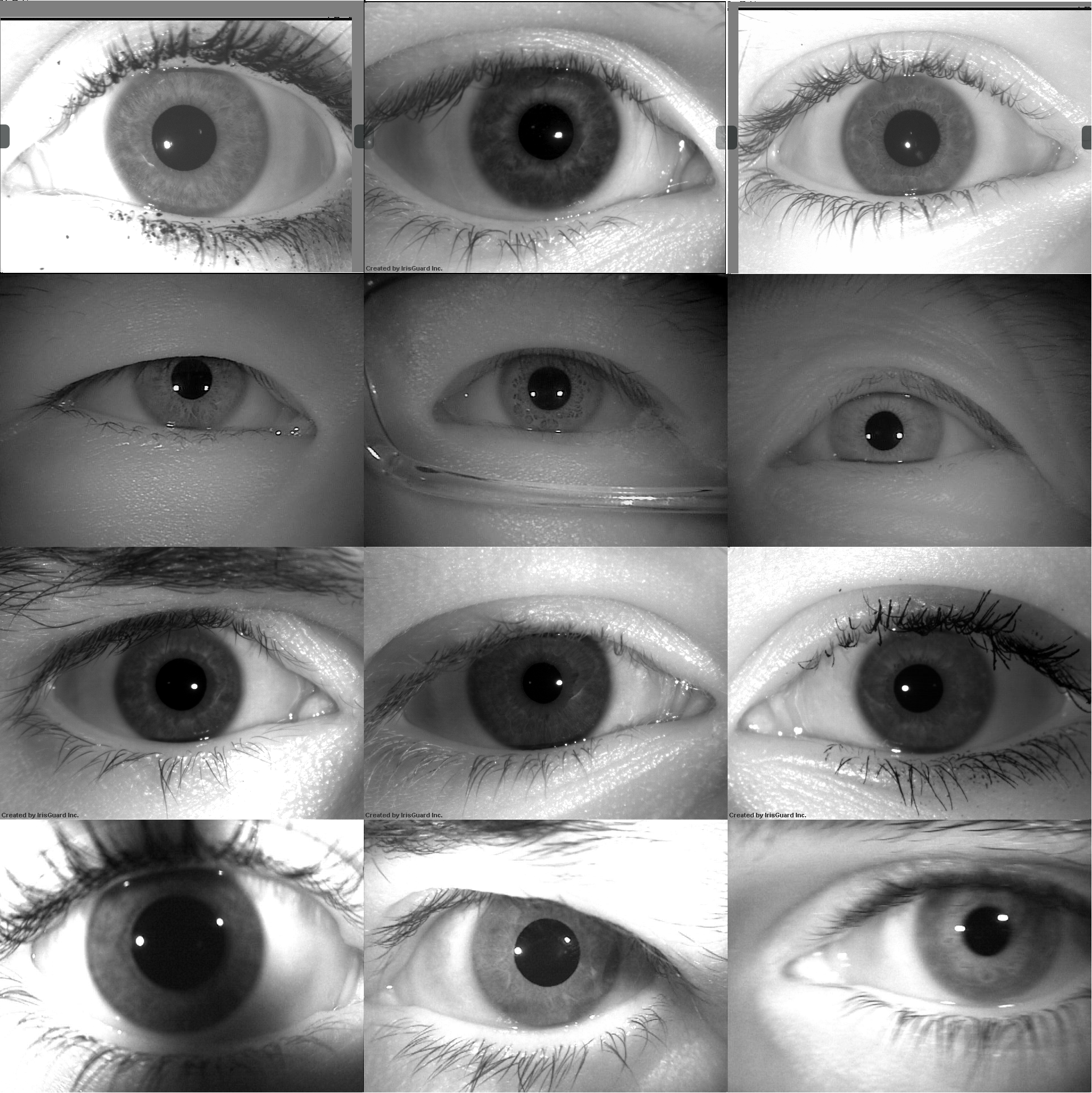}
    \caption{Top row: Examples from the network training and validation sets. Second row: CASIA-Iris-Thousand. Third Row: Warsaw Subset from LivDet-Iris 2017. Bottom Row: WVU Non-Ideal Iris Database-Release 1.}
    \label{databases}
\end{figure}

Figure \ref{databases} shows three examples from each of the databases used in this paper. The top row represents the data used to train the network and the bottom three rows are samples from each of the three subject-disjoint testing datasets.
\label{sec:database}

\subsubsection{Training, Validation and Testing Datasets}
In this work a set of in-house data was employed to train the network. This set consists of 339,400 iris images from 3,000 classes. All images in this dataset are live irises and are of size $640\times480$. Images in this set were acquired using the LG 2200, LG 4000 and IrisGuard AD100 sensors. In the training pipeline, this data is further subdivided into training and validation. The training subset consists of 80\% of the data and the validation set is the remaining 20\%. The training and validation subsets are subject-disjoint meaning out of the 3,000 classes, 2,400 are used in training and 600 for validation. %To simplify explanations, this dataset will be referred to as the \textit{network training set}.

To test the trained network, it was decided to use the following three subject-disjoint and cross-sensor databases:

\paragraph{CASIA-Iris-Thousand V4 \cite{casia-database}} \, which contains 20,000 images from 1,000 subjects. Both left and right eye images were acquired for each subject, meaning there are 2,000 classes total in this dataset. All images in this dataset were acquired using the IKEMB-100 sensor from IrisKing.

\paragraph{WVU Non-Ideal Iris Database-Release 1 \cite{crihalmeanu2007protocol}} , which contains 3,042 live iris images from a total of 231 subjects, totalling 461 individual classes. Included in this database are images collected in real environments including samples of off-angled, blurred, sensor noise and occlusions. 

\paragraph{LivDet 2017 Liveness Detection-Iris -- Warsaw Subset (live subset) \cite{yambay2017livdet}} , which consists of 5,168 images from 468 classes. The sensor used to capture these images was the IrisGuard AD100 sensor and a setup composed of Aritech ARX-3M3C camera with SONY EX-View CCD sensor (with an increased NIR sensitivity), equipped with Fujinon DV10X7.5A-SA2 lens and B+W 092 NIR filter. This database contains clean, constrained, high-quality images. 

 \begin{figure*}[t]
    \centering
    \includegraphics[width=1\textwidth]{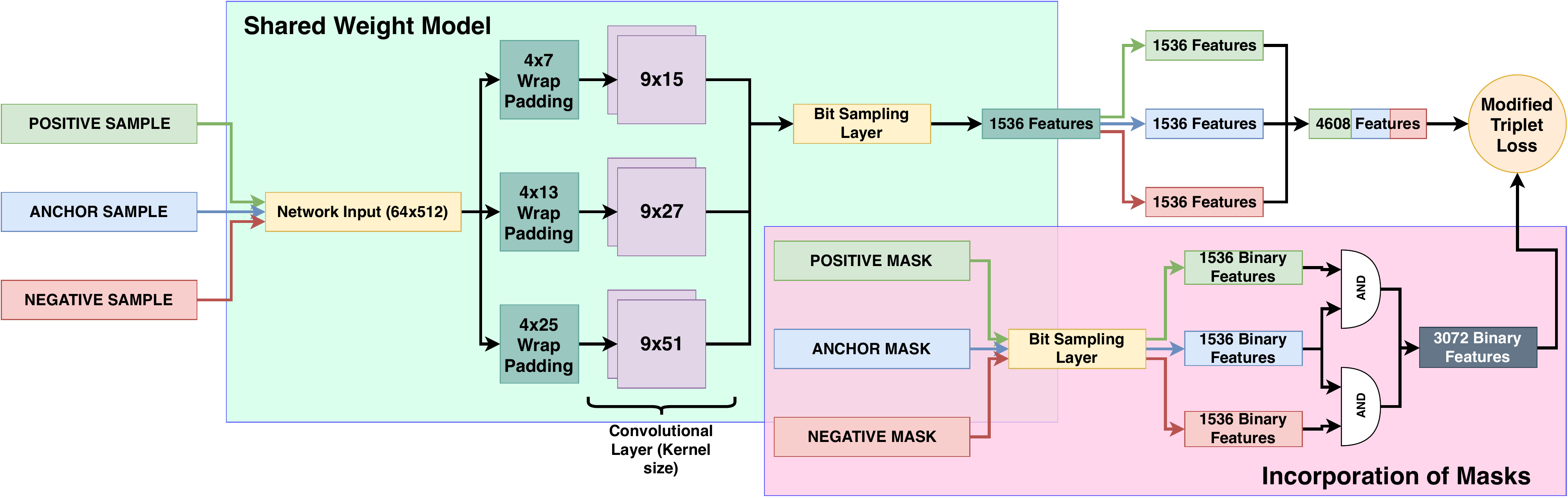}
    \caption{System Overview: On the left we see three inputs: a positive sample, an anchor sample and a negative sample. This outlines the use of triplet loss. Each image is then passed through a shared weight model to produce a feature vector of size $1536$ using a single convolutional layer and a bit sampling layer. Images are padded with actual iris data. These three feature vectors from the input images are then concatenated and passed to the loss function. Occlusion masks are also incorporated in the loss so that regions that do not correspond to iris texture are not included in the calculations.}
    \label{overview}
\end{figure*}

\subsubsection{Genuine and Impostor List Generation}

The genuine list for both CASIA, Warsaw and WVU databases includes all possible genuine pairs. For the impostor list generation, in order to reduce the number of impostor comparisons but also maintain a representative sample of comparisons, a sample selection strategy was devised. This strategy works as follows: firstly, we only compare images of right eyes with other classes of right eyes and images of left eyes with classes of left eyes. Next, for each class, we select one random image from this reference class and compare it to a random image from another impostor class of the same side of the face. %% IS SIDE A GOOD WORD HERE?
We then select another random image from the first reference class and compare with another different impostor class of the same eye. This is repeated until every class has a comparison with an image in every other class of the same eye type.

For the CASIA-Iris-Thousand database there are 2,000 classes consisting of 1,000 of both left and right eyes. Each of these classes contains 10 images, totalling to 20,000 images. For genuine comparisons, this means there are $10 \choose 2$ $= 45$ genuine comparisons per class, resulting in 90,000 total genuine comparisons. For impostor comparisons, using the above described method results in $999$ comparisons for each class (one class to  all other classes of eyes on the same side of the face). This results in $2,000 \times 999$ $= 1,998,000$ impostor comparisons.

Applying the same methodology, the resulting number of genuine comparisons for the Warsaw database is 47,408 and the number of impostor comparisons equals 218,556. For the WVU database there are 10,344 genuine comparisons and 212,060 impostor comparisons.

\subsection{Image Preprocessing}

There were two steps undertaken to preprocess the images before they could be used in the network. The first preprocessing step was to segment and normalize the iris to a standard size and after this data alignment was performed.

\subsubsection{Segmentation and Normalization}

The software employed to segment and normalize all iris images in this work is OSIRIS \cite{othman2016osiris}. The open-source OSIRIS software outputs a normalized iris image of size $64\times512$, as well as the corresponding normalized occlusion mask of the same size. During image segmentation and normalization, if there was a failure, this sample was excluded from the subset. The reasoning for excluding these samples is that only valid training samples are desired so that the network learns iris-related features exclusively. On the network training set there was only 101 failures out of the 339,400 images, so the final network training set comprises 339,299 normalized iris images and occlusion masks from 3,000 classes. For the CASIA-Iris-Thousand database, there was 30 failures and a final testing dataset includes 19,970 images from 2,000 classes. For the Warsaw database, out of the 5,168 images there were 13 failures and for the WVU database out of the 3,043 images there were 3 failures.

\subsubsection{Data Alignment}

To prevent contradictory inter-class information from being presented to the network, a data alignment scheme was developed and hence incorporates rotation compensation \cite{Daugman_PAMI_1993} into the iris recognition system. The data alignment method operates as follows: for each unique class, the image with the highest mean Pearson Correlation Coefficient (PCC) when compared to all images in that same class is selected as the reference. Using this reference image, each of the other images in the same class are individually rotated one pixel at a time along the x-axis until they reach the point with the highest positive PCC in relation to the reference image. The corresponding occlusion masks are also aligned accordingly.

\subsection{Open-source Gabor Kernels}

OSIRIS software offers six hand-crafted Gabor kernels. These Gabor kernels consist of three pairs at different scales: $9 \times 15$, $9 \times 27$ and $9 \times 51$. Each image is convolved with each of the six kernels independently and then from each of the six feature outputs 256 points are extracted from predefined coordinates. This results in an Iris Code of length 1536. The list of which 256 coordinates to extract the value from is included with the OSIRIS tool, and identical bit sampling was applied in our network.

\subsection{Network Architecture}

\textit{One Convolutional Layer:} For this work the network architecture was designed to implement a Daugman-style approach to iris recognition, as shown in Figure \ref{overview}. To this end, the model only contains one single convolutional layer with six feature maps (three pairs of filters at different scales of $9 \times 15$, $9 \times 27$ and $9 \times 51$ to mimic the filters provided with the OSIRIS software). %% TODO: rephrase
The \textit{sigmoid} activation function is applied to the output of this convolutional layer to normalize all features to the $[0,1]$ space.
 
The input to this network is a normalized iris image of size $64 \times 512$. The associated normalized mask is used in the loss function and as such is not regarded as the input to the network.
After this convolutional layer, the output of each feature map is concatenated to form a feature vector of size $64 \times 512 \times 6$. 
 
\textit{Bit Sampling Layer:} Upon concatenation of the convolutional outputs, this $64 \times 512 \times 6$ feature vector is passed into an non-trainable custom layer we denote as the \textit{bit sampling layer}. The purpose is to extract final iris code bits from the predefined 256 locations in each of the six resulting feature maps, as provided by the OSIRIS tool, resulting in a feature vector of size $1,536$ in the range of 0 to 1. This feature vector serves as the output of the network.
 
\textit{Wrap Padding:}
To prevent the shrinking of the input image after undergoing convolution and to make sure that the selected bits are not just representing a region of zero padding, the input images are padded using a wrapping technique before undergoing convolution. This padding takes one side of an image and wraps it around to the opposite side of varying size depending on the convolutional filter size. As shown in Figure \ref{overview}, for the $9 \times 15$ filter, a pad of 4 pixels in both y-directions and 7 pixels in each x-direction is applied. % After the introduction of padding, the result of the convolution is a $64 \times 512$ feature map that does not have any occlusions due to convolutional shrinking. 
%  In order to replicate the Daugman approach to iris recognition as closely as possible, the network architecture used in this work is only one convolutional layer deep. This convolutional layer is split up into three separate pairs of filter sizes, totalling 6 feature maps. This

\subsection{Triplet Loss}

Since the problem of developing effective kernels for meaningful iris feature extraction is inherently a verification problem rather than a classification problem, the application of a triplet loss seemed natural. The goal of triplet loss is to minimize the feature embedding distance between instances of the same class and to maximise the feature embedding distance of different classes. To do this triplets are formed, which include an image called a \textit{anchor}, another image of the same class as the anchor called the \textit{positive} and an image of a different class to the anchor and positive called the \textit{negative}. The triplet loss then takes the output of the network \ie a feature vector for each of the anchor, positive and negative and then compares these feature vectors. Based on these vectors the network weights are updated to push the seen negative classes further and further away from the anchors while pulling the instances of the same class closer and closer.

\subsubsection{Network Modifications}

To extend our single convolutional layer model to be used in a network with triplet loss, we instantiate this network with shared weights for three inputs: the anchor, the positive and the negative, as shown in Fig. \ref{overview}. The output vector of size $1536$ for each input is then concatenated to form a feature vector of $4608$ where the first $1536$ elements correspond to the anchor, the second $1536$ elements correspond to the positive and the final $1536$ elements represent the negative image. This concatenated feature vector appears in the loss function as the predicted label and from here we extract each image's features and calculate the distance.

\subsubsection{Incorporation of Masks}

In iris recognition, it is essential that occlusion masks are used in iris code comparisons. These occluded regions give us no useful information about the iris and also may appear very similar across different subjects. To this end, we propose a methodology to incorporate the exclusion of these occluded regions into the loss function.

For both the anchor/positive and anchor/negative pairs, a combined occlusion mask is calculated by excluding each pixel that is occluded in either the anchor or the positive/negative. The result is two masks of size $64 \times 512$ that only represents regions that contain useful iris information in both individual samples for both pairs. We then pass these masks through the same \textit{bit sampling layer} defined above to output the bits in the mask that represent the points that were sampled from the output of the convolution, resulting in two binary vectors of length $1536$. 

To utilize the fact that when using triplet loss there is no true labels for the data as the function is simply trying to maximise a distance between samples rather than between the true and predicted labels, true labels can be defined to anything the user wishes. Using this knowledge, in order to make the combined masks of both the anchor/positive and anchor/negative pairs appear in the loss function, the true labels are set to be a concatenation of these two binary vectors, the first $1536$ features corresponding to the combined mask of the anchor and positive samples and the second $1536$ features corresponding to the combined mask of the anchor and negative samples. Because the true and predicted vectors must be the same size, we add a vector of zeros of size $1536$ to the concatenation of the two combined masks. This results in a total feature vector of $4608$, the same as the predicted vector detailed above. This final set of zeros of length $1536$ is simply discarded in the loss function.

\begin{figure*}[t]
  \begin{subfigure}[b]{1\linewidth}
      \begin{subfigure}[b]{0.32\linewidth}
          \centering
          \includegraphics[width=1\columnwidth]{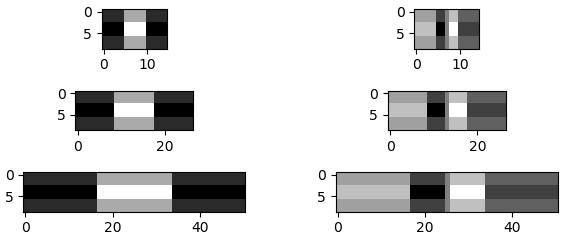}
      \end{subfigure}
      \vspace{0.3em}
      \hfill
      \begin{subfigure}[b]{0.32\linewidth}
          \centering
          \includegraphics[width=1\columnwidth]{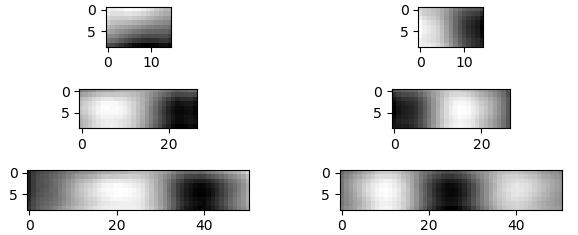}
      \end{subfigure}
      \vspace{0.3em}
      \hfill
      \begin{subfigure}[b]{0.32\linewidth}
          \centering
          \includegraphics[width=1\columnwidth]{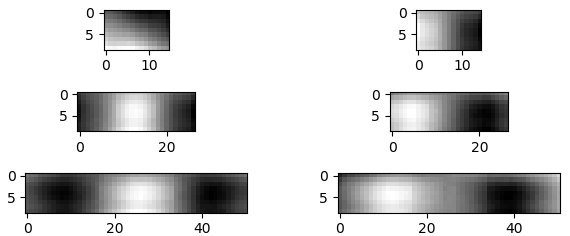}
      \end{subfigure}
      \vspace{0.3em}
  \end{subfigure}
  \begin{subfigure}[b]{1\linewidth}
      \begin{subfigure}[b]{0.327\linewidth}
          \centering
          \includegraphics[width=1\columnwidth]{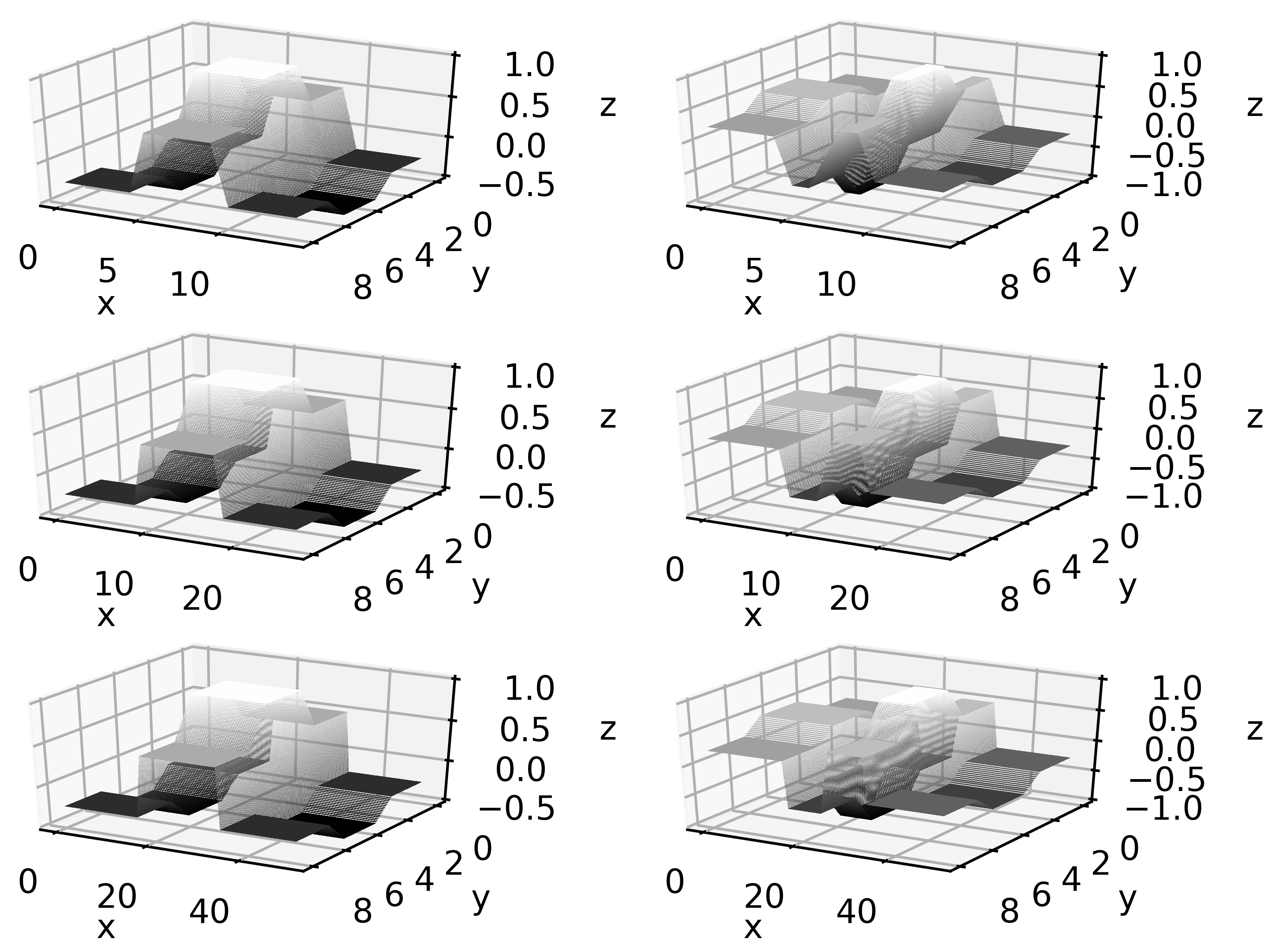}
          \caption{Original Gabor Kernels}
      \end{subfigure}
      \hfill
      \begin{subfigure}[b]{0.327\linewidth}
          \centering
          \includegraphics[width=1\columnwidth]{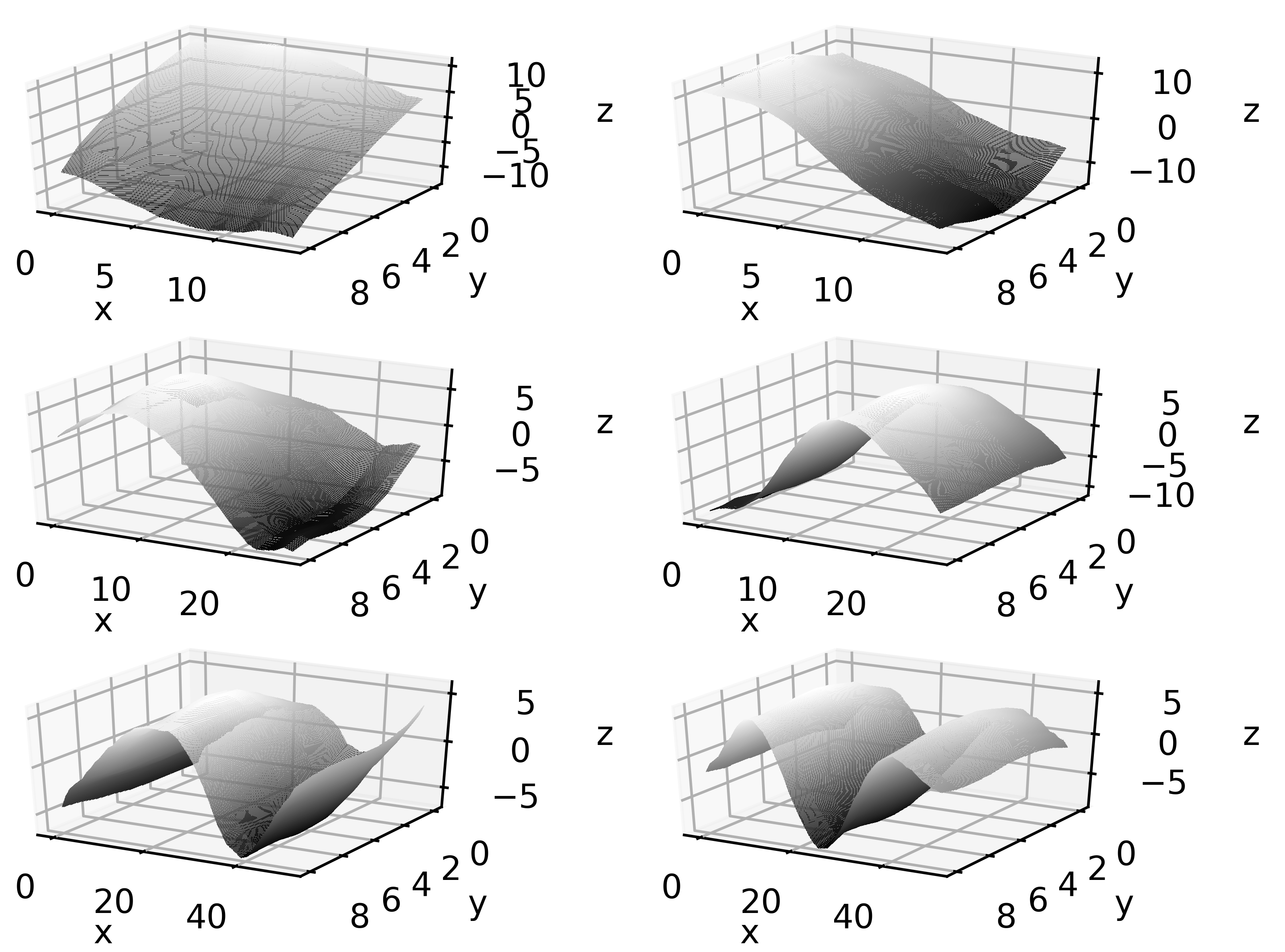}
          \caption{Randomly Initialized Kernels}
      \end{subfigure}
      \hfill
      \begin{subfigure}[b]{0.327\linewidth}
          \centering
          \includegraphics[width=1\columnwidth]{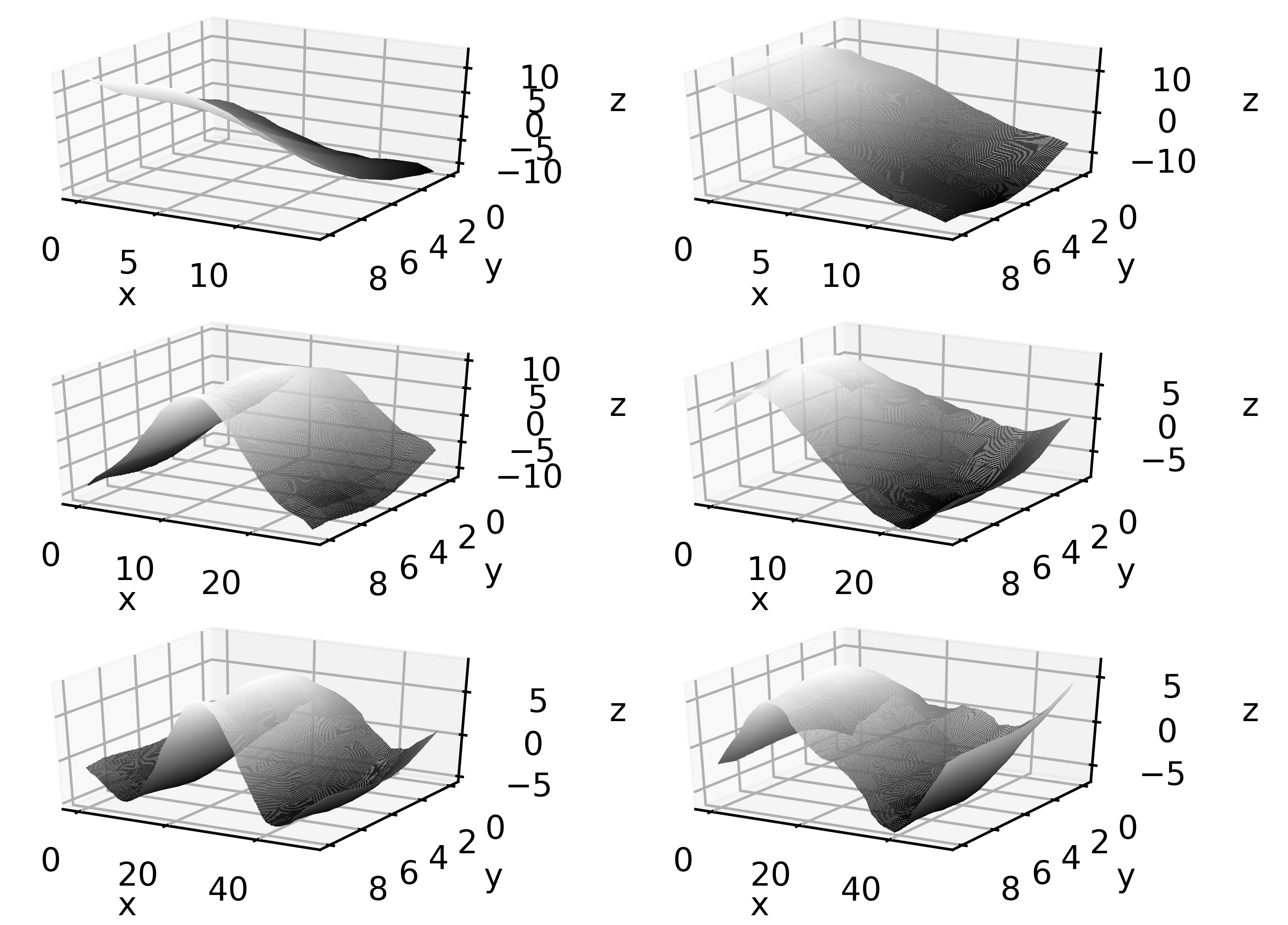}
          \caption{Gabor Initialized Kernels}
      \end{subfigure}
  \end{subfigure}
  \caption{Visualization of all kernels used in this work. The top three rows represent the 2D version of the bottom three rows of 3D visualizations. The leftmost two columns represent the open-source Gabor kernels, the middle two rows are the kernels learned using randomly initialized weights and the rightmost two columns are the kernels learned using Gabor initialized weights.}
  \label{kernels}
\end{figure*}

\subsubsection{Distance Metric}

In general, for triplet loss based solutions, the Euclidean squared distance between two samples is used as the distance metric \cite{ahmad2019thirdeye}. For the task of iris recognition, however, we use a metric that incorporates occlusion masks: 

$$
d = \frac{\sum_{i=0}^{N-1} \big(\abs{s_1(i) - s_2(i)} m_1(i)  m_2(i)\big)}{\sum_{i=0}^{N-1}\big(m_1(i) m_2(i)\big)}
$$

where $s_{1}$ and $s_{2}$ denote feature vectors of two samples being compared, $m_{1}$ and $m_{2}$ denote the corresponding masks (one denotes iris pixel, zero denotes occlusion), and $N=1536$. One may see a close similarity between $d$ and the fractional Hamming distance calculated when $s_{1}$ and $s_{2}$ are binary vectors. Ideally, the value of $d$ would be zero for the anchor/positive pair ($d_{ap}$) and one for the anchor/negative pair ($d_{an}$).

%The Hamming distance calculation sums up how many bits in a pair of binary iris codes are different divided by the total length of the iris codes, whereas because we do not have binary features naturally, we try to push these features to be binary and therefore make the calculation similar to hamming distance. 

\subsubsection{Soft-margin} % see third eye paper

Traditional triplet losses use a margin value $\alpha$ to specify how far we wish to separate out the anchor/positve ($d_{ap}$) and anchor/negative ($d_{an}$) as follows: 
\[loss =   max(0, d_{ap} - d_{an} + \alpha)\]

However, setting $\alpha$ is dataset-specific and requires hyper-parameter tuning to find out what works best as this value affects training accuracy and convergence. Additionally, this means the loss calculation acts as a hinge function. Instead, we applied a soft margin, as proposed in \cite{hermans2017defense}: 
\[loss =   log(1 + exp(d_{ap} - d_{an}))\]

This soft-margin loss function acts similarly to the hinge function but instead of having a hard cut-off, the loss decays exponentially and also does not require the discovery of an appropriate value of $\alpha$.

\begin{figure*}[t]
  \begin{subfigure}[b]{1\linewidth}
      \begin{subfigure}[b]{0.32\linewidth}
          \centering
          \includegraphics[width=1\columnwidth]{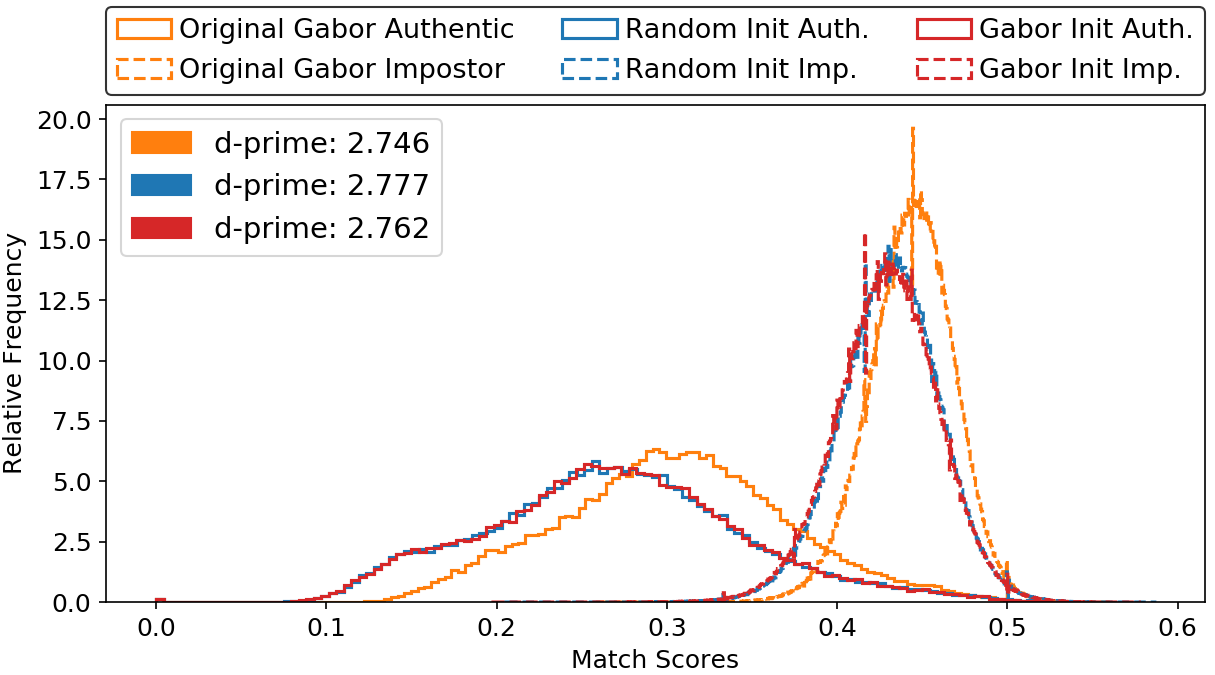}
      \end{subfigure}
      \vspace{0.3em}
      \hfill
      \begin{subfigure}[b]{0.32\linewidth}
          \centering
          \includegraphics[width=1\columnwidth]{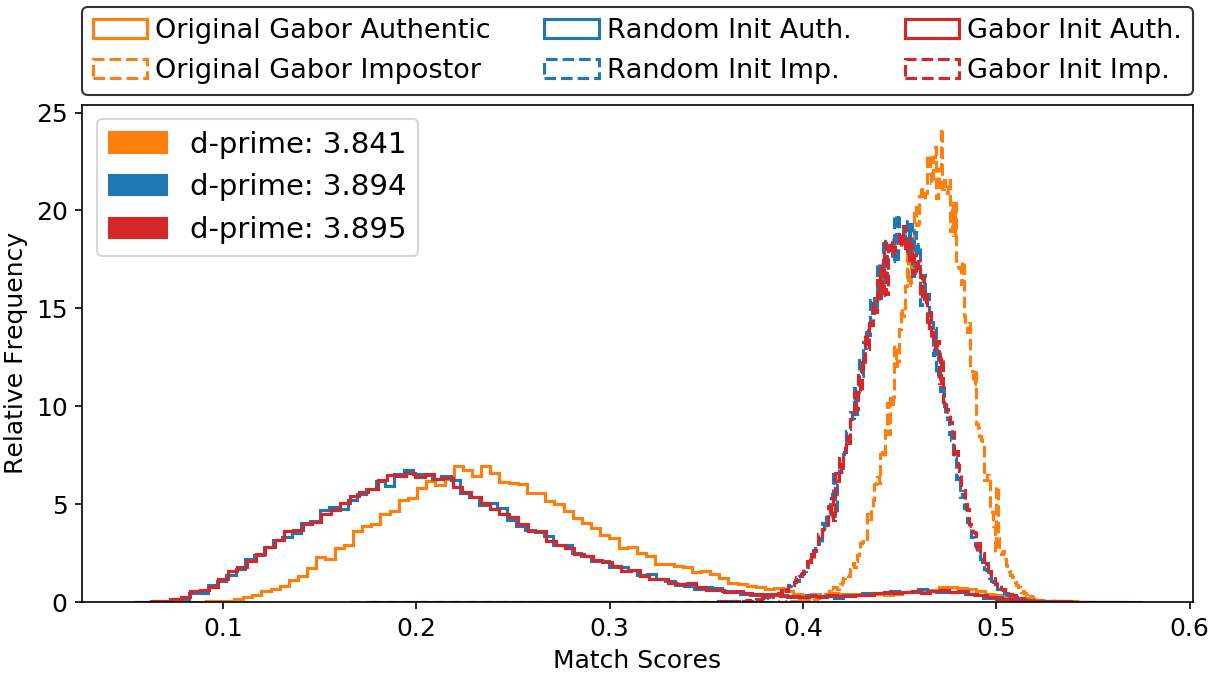}
      \end{subfigure}
      \vspace{0.3em}
      \hfill
      \begin{subfigure}[b]{0.32\linewidth}
          \centering
          \includegraphics[width=1\columnwidth]{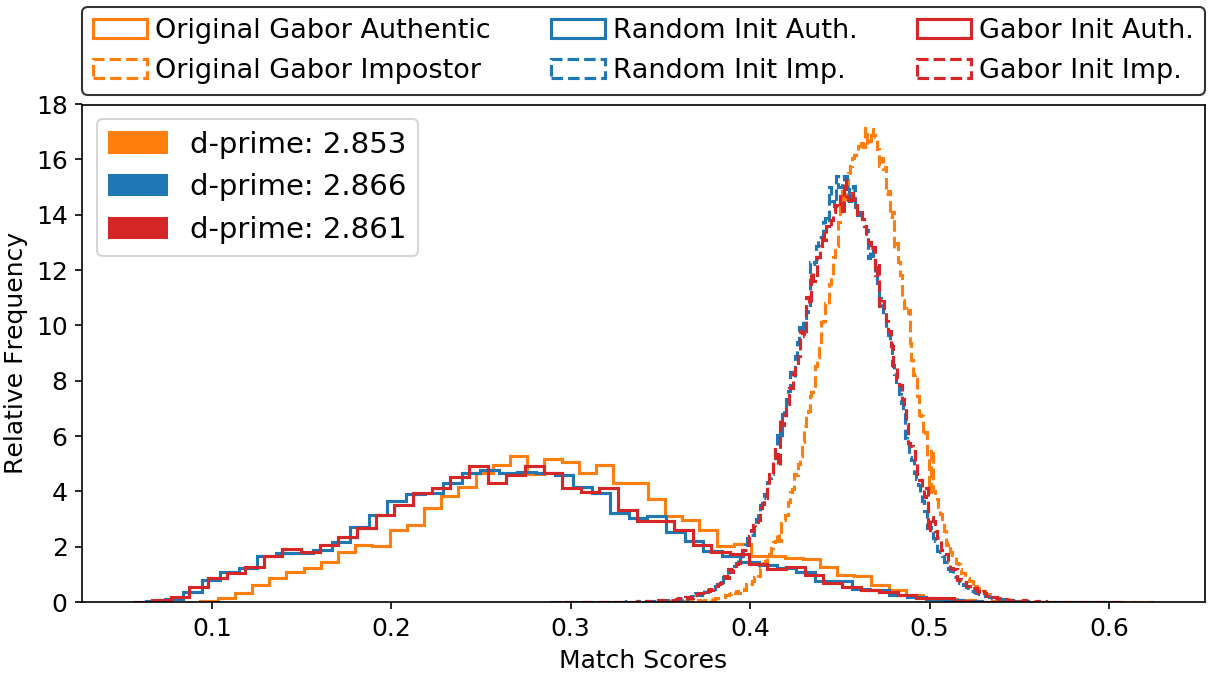}
      \end{subfigure}
      \vspace{0.3em}
  \end{subfigure}
  \begin{subfigure}[b]{1\linewidth}
      \begin{subfigure}[b]{0.327\linewidth}
          \centering
          \includegraphics[width=1\columnwidth]{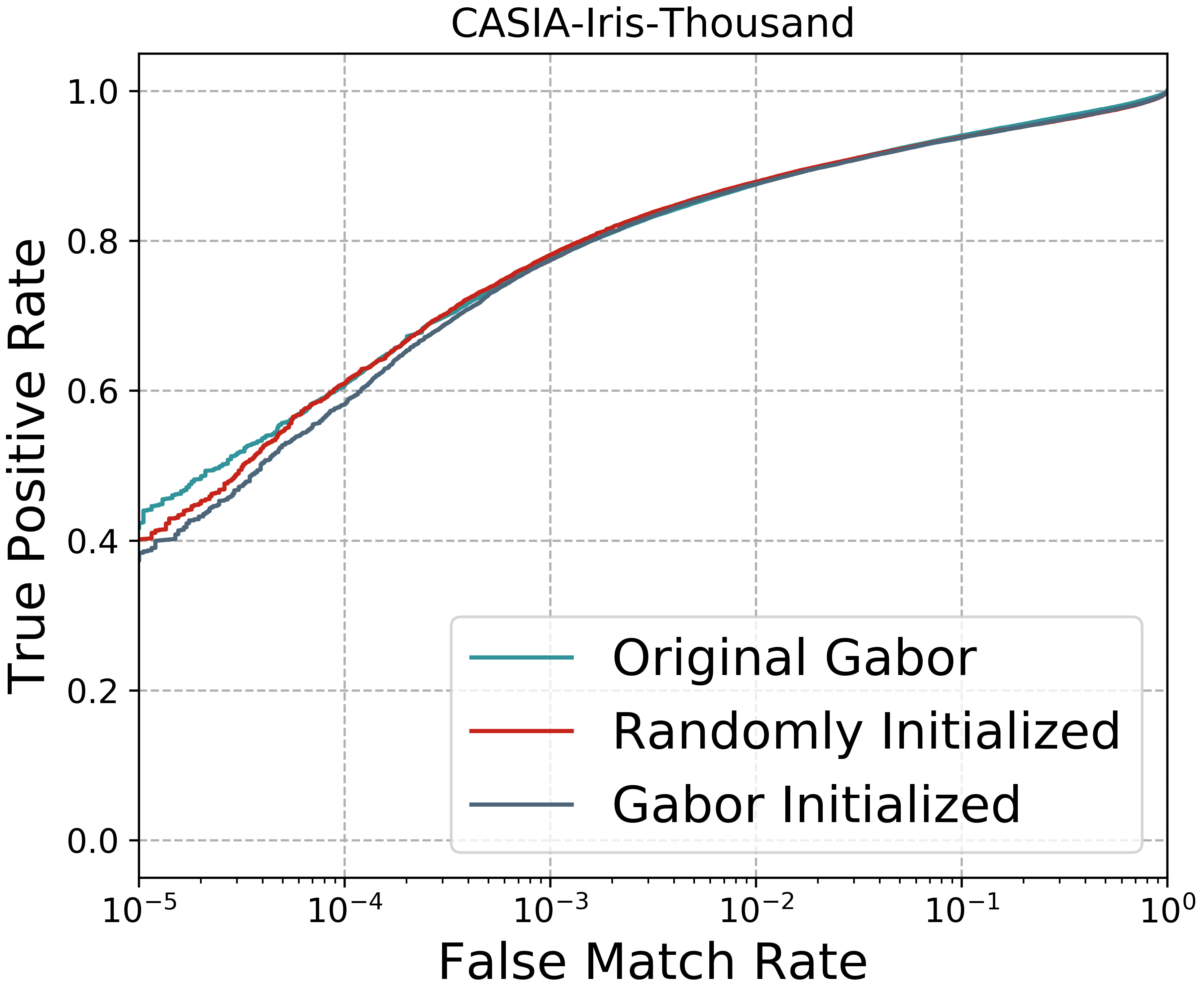}
          \caption{CASIA-Iris-Thousand}
      \end{subfigure}
      \hfill
      \begin{subfigure}[b]{0.327\linewidth}
          \centering
          \includegraphics[width=1\columnwidth]{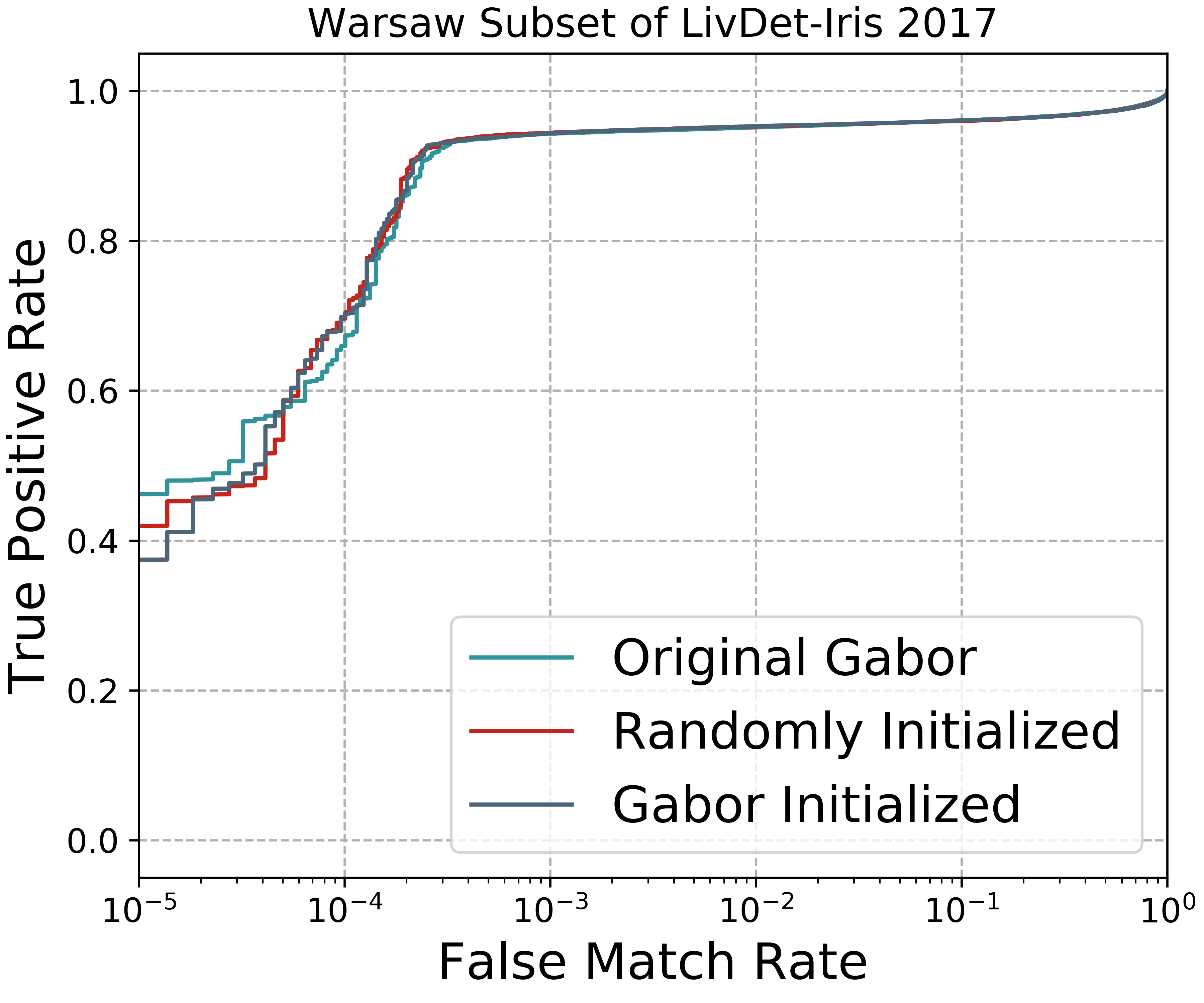}
          \caption{Warsaw}
      \end{subfigure}
      \hfill
      \begin{subfigure}[b]{0.327\linewidth}
          \centering
          \includegraphics[width=1\columnwidth]{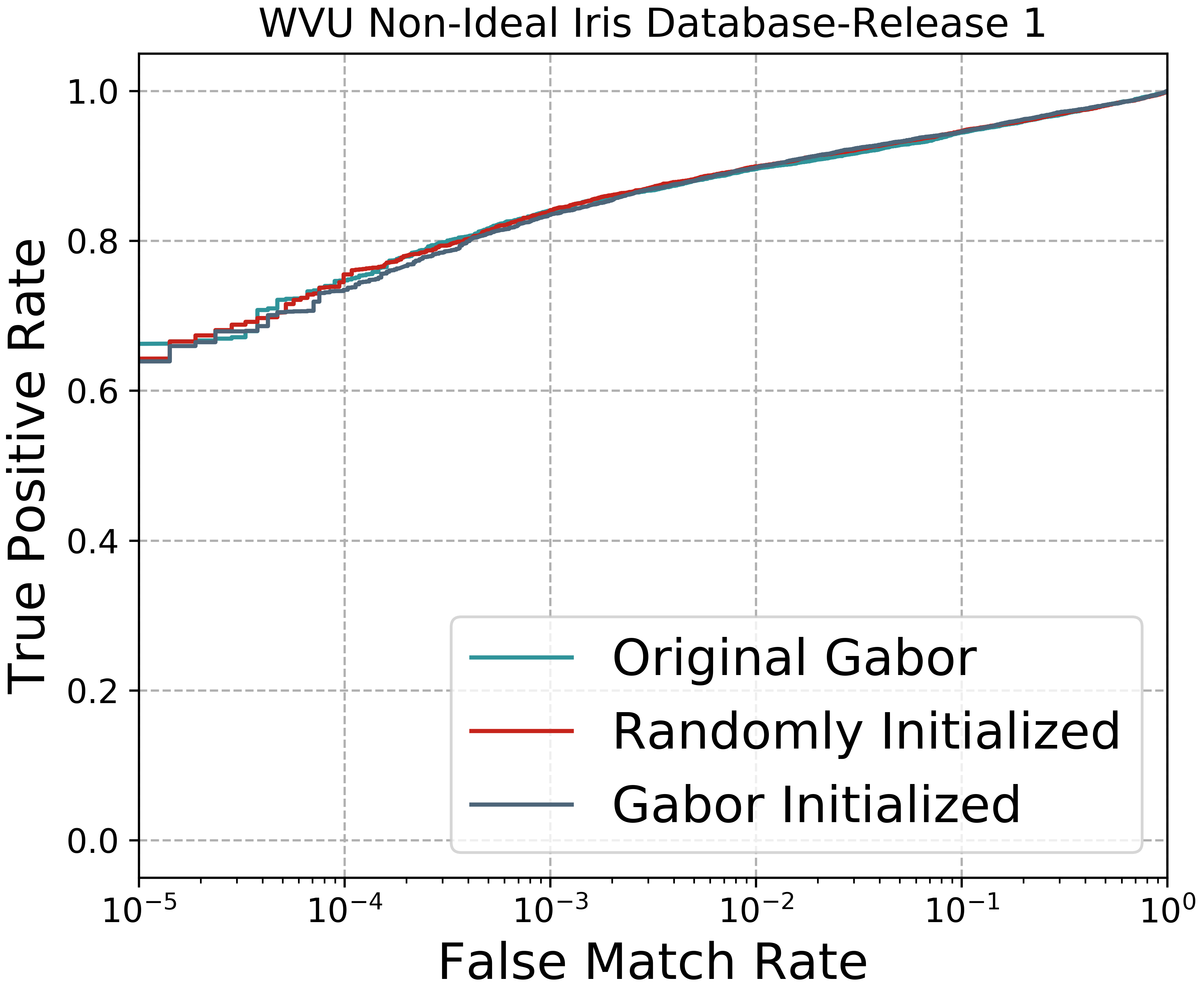}
          \caption{WVU}
      \end{subfigure}
  \end{subfigure}
  \caption{Results for all three testing databases including the genuine/impostor distributions and the ROC curves. {\bf Left column:} CASIA-Iris-Thousand. {\bf Middle column:} Warsaw Subset of LivDet-Iris 2017. {\bf Right column:} WVU Non-Ideal Iris Database-Release 1.}
  \label{fig:roc}
\end{figure*}

\subsubsection{Triplet Mining} %%% Very important person re-id paper

It has been demonstrated that the use of triplet mining has shown an increase in training accuracy and generalization \cite{hermans2017defense}. Following this, it was decided to implement the selection of \textit{batch hard} triplets for training. If the batch size is $X$, batch generation is completed by randomly selecting $X$ unique classes (this set of classes is denoted as $B$). One random image is then selected from each class in $B$ as the anchor and another random image from the same class is selected as the positive. We do not do any mining on the anchor/positive pair. When selecting the negative for a triplet, for each of the anchor/positive pairs independently we randomly select $X$ new classes that do not appear in $B$ (denoted as $B'$). One random image is then selected from each class in $B'$. The embeddings are then calculated using the current model weights when setting each of these images from $B'$ are set as the negative for the current anchor/positive sample. Distances $d_{ap}$ and $d_{an}$ are then calculated as described above. The negative that corresponds to the smallest distance is set as the negative for that triplet. 

This is repeated for each anchor/positive pair in $B$. It is important to note that occlusion masks were taken into account when calculating $d_{an}$. After this batch is used in training, the model weights will update such that the hardest samples \ie $d_{an}$ gradually grows further away from $d_{ap}$. This process of selecting the hardest negative only is referred to as \textit{batch hard negative mining}. Because this hard mining is only performed on a small sample of the overall data, these discovered hardest triplets can be considered \textit{moderately hard} with respect to the entire dataset. It was found that a batch size of $64$ was optimal with regards to time. Since this operation of batch hard mining is $O(n^{2})$, increasing this batch size beyond $64$ results in batches taking too long to be created for a feasible training regimen.

Due to GPU memory and time constraints, from the within validation subset we use $2048$ random triplets to act as the actual validation set for the training. This validation set is persistent throughout training so that an accurate representation on training progression is provided. Training was stopped for both the randomly initialized weight network and gabor initialized network after 20,000 batches as this was subjectively decided to represent the convergence of both networks based on the validation loss.

% \subsection{Stopping Criteria}

% If the validation loss does not increase over two validation steps the training ends. Each validation step occurs every 100 batches. The network that was initialized randomly converged after 6300 batches whereas the network that was initialized from Gabor kernels converged after 3900 batches. This hints at the fact that training start point may influence convergence speed.

\section{Evaluation}

\subsection{Kernel Analysis}

Figure \ref{kernels} shows 2D and 3D plots of the used kernel sets: (a) original approximations of Gabor kernels as offered by the existing open-source tool, (b) kernels that were learned from data using random weight initialization, and (c) kernels learned from the same data using Gabor initialized weights. From this figure it is clear that the data-driven kernels did not converge on a similar structure to the original Gabor kernels, instead, some interesting behaviour is observed. 

On initial observation of the data-driven kernels, both sets appear to have converged to similar structures. For both pairs of data-driven $9 \times 15$ kernels, the network converged on what appears to be edge detectors. This can be seen in the 3D plots in that the kernels look like flat planes, meaning they extract edge based features. For the $9 \times 27$ kernels, the network converged on blob detectors and for the $9 \times 51$ kernels the final kernels look like simple waves. There is slight differences between the randomly initialized kernels and the Gabor initialized kernels, however, both sets converged on similar looking kernels even though the weights were initialized completely differently. Both the $9 \times 15$ and the $9 \times 27$ data-driven pairs do not display Gabor wavelet properties, so we can say these converged to something deviant of Gabor. The $9 \times 51$ kernels for both weight sets, however, do exhibit some Gabor wavelet properties. %, although are not strictly Gabor. % They have converged to a wave which can be interpreted as a simple Gabor wavelet, however, they could also be interpreted as a Sine or Cosine wave. %% THIS OKAY?

\subsection{Evaluation Methodology}

To perform a fair comparison between the data-driven kernels and the open-source Gabor kernels, the testing methodology involved the use of the OSIRIS tool to generate the matching scores. First, segmentation and normalization is performed on the testing databases using the out-of-the-box OSIRIS configuration. Then matching scores are calculated using the iris codes generated by original Gabor kernels. Then, we swap out these original kernels and replace them with the kernels that were learned from data and then the exact same encoding and matching as before is performed. We make sure that the learned kernels all sum to zero by subtracting the mean value of each kernel on all elements in that kernel, as OSIRIS binarizes the filter responses at zero to produce the iris code.

\subsection{Hand-crafted versus Data-driven}

Figure \ref{fig:roc} shows genuine/impostor distributions as well as the corresponding ROC curves, along with the decidability value $d'$, which evaluates the separation between two distributions for each of the three testing datasets. The same methodology is applied to the original Gabor kernels, the randomly initialized data-driven kernels and the Gabor initialized data-driven kernels. These experiments are conducted independently on all three testing datasets.

From the results in Figure \ref{fig:roc}, the performance of the learned kernels in comparison to the hand-crafted Gabor kernels is very similar, with the hand-crafted set and the data-driven kernels performing almost identically on all three testing databases. These hand-crafted Gabor kernels were developed for the purpose of extracting features on all iris images. The data-driven kernels were learned on a large database of iris images. Even though both the hand-crafted and data-driven kernels both extract different features, it seems that both perform comparably in extracting unique iris features. The interesting takeaway from this is what the kernels actually converged to. The data-driven kernels did not converge to Gabor kernels, which hints that an optimal solution for the kernels for iris recognition may be from outside the family of Gabor wavelets. If a simple network with minimal parameters can learn something that performs almost identically to the time-intensive task of hand-crafting these Gabor kernels, it may be that there are some undiscovered kernel set that is optimal. We were asking ourselves in the introduction whether a neural network that replicates a Daugman approach to iris recognition `thinks' that Gabor kernels are optimal by converging on this family of kernels, however, we do not see this!

\subsection{Weight Initialization}

Another question posed in the introduction was whether the weight initialization for this shallow network has an affect on accuracy. To do this we trained the network from two starting points, one using random weight initialization and the other using the hand-crafted Gabor kernels as the starting point. From Figure \ref{fig:roc} we can see that there is no discernible difference between these kernels sets in terms of results, even though it can be seen in Figure \ref{kernels} that these two sets of data-driven kernels appear significantly different than open-source Gabor. There are some clear similarities between the kernels the network converged on, but none of these are explicitly Gabor. This means that no matter the starting point, and even if the network is started from Gabor, it does not converge to Gabor instead opting for a combination of edge detectors, blob detectors and simple waves.

\section{Discussions and Conclusions}

In this work we develop and release a neural network framework that mimics Daugman's approach to iris recognition that can be used to learn data-driven kernels for the purpose of iris recognition. We use this framework along with a large iris database to learn data-driven kernels and then we compare these learned kernels to hand-crafted, open-source Gabor kernels included with the OSIRIS tool. 

To answer the questions posed in the introduction, we see that the learned kernels perform almost identically to the hand-crafted kernels. Shown is that the initial weight initialization also does not play a huge role in the convergence of the network as the two weight sets converged on similar looking and performing kernels. 

A goal of this paper was to determine if a neural network that mimics Daugman's approach converges on Gabor kernels, widely accepted as the optimal kernel set for iris recognition. Through learning data-driven kernels we show that this may not be true. The data-driven kernels learned in this work do not resemble Gabor, instead are comprised of a mixture of edge detectors, blob detectors and simple waves. This may mean that the optimal kernel set for iris recognition could be from a broader family than Gabor kernels alone. 

Through experimentation, we see that the learned kernels perform very similarly to the hand-crafted kernels. This could mean that we have hit the maximum accuracy possible using a one convolutional layer with six feature maps. To increase accuracy, it seems that additional operations are needed such as the addition of extra layers in the network. 

{\small
\bibliographystyle{ieee}
\bibliography{egbib}

\begin{thebibliography}{10}\itemsep=-1pt

\bibitem{casia-database}
Chinese academy of sciences institute of automation.
\newblock http://biometrics.idealtest.org/dbDetailForUser.do?id=4.
\newblock Accessed: 12-15-2019.

\bibitem{aboyd-github}
Repository of supplementary material.
\newblock Link removed to preserve anonymity.
\newblock Accessed: 12-15-2019.

\bibitem{ahmad2019thirdeye}
S.~Ahmad and B.~Fuller.
\newblock Thirdeye: Triplet based iris recognition without normalization.
\newblock {\em arXiv preprint arXiv:1907.06147}, 2019.

\bibitem{crihalmeanu2007protocol}
S.~Crihalmeanu, A.~Ross, S.~Schuckers, and L.~Hornak.
\newblock A protocol for multibiometric data acquisition, storage and
  dissemination.
\newblock In {\em Technical Report, WVU, Lane Department of Computer Science
  and Electrical Engineering}. 2007.

\bibitem{daugman2015information}
J.~Daugman.
\newblock Information theory and the iriscode.
\newblock {\em IEEE transactions on information forensics and security},
  11(2):400--409, 2015.

\bibitem{Daugman_PAMI_1993}
J.~G. Daugman.
\newblock High confidence visual recognition of persons by a test of
  statistical independence.
\newblock {\em {IEEE Transactions on Pattern Analysis and Machine
  Intelligence}}, 15(11):1148--1161, November 1993.

\bibitem{gangwar2016deepirisnet}
A.~Gangwar and A.~Joshi.
\newblock Deepirisnet: Deep iris representation with applications in iris
  recognition and cross-sensor iris recognition.
\newblock In {\em 2016 IEEE International Conference on Image Processing
  (ICIP)}, pages 2301--2305. IEEE, 2016.

\bibitem{hermans2017defense}
A.~Hermans, L.~Beyer, and B.~Leibe.
\newblock In defense of the triplet loss for person re-identification.
\newblock {\em arXiv preprint arXiv:1703.07737}, 2017.

\bibitem{jones1987evaluation}
J.~P. Jones and L.~A. Palmer.
\newblock An evaluation of the two-dimensional gabor filter model of simple
  receptive fields in cat striate cortex.
\newblock {\em Journal of neurophysiology}, 58(6):1233--1258, 1987.

\bibitem{othman2016osiris}
N.~Othman, B.~Dorizzi, and S.~Garcia-Salicetti.
\newblock Osiris: An open source iris recognition software.
\newblock {\em Pattern Recognition Letters}, 82:124--131, 2016.

\bibitem{wang2019towards}
K.~Wang and A.~Kumar.
\newblock Towards more accurate iris recognition using dilated residual
  features.
\newblock {\em IEEE Transactions on Information Forensics and Security}, 2019.

\bibitem{yambay2017livdet}
D.~Yambay, B.~Becker, N.~Kohli, D.~Yadav, A.~Czajka, K.~W. Bowyer,
  S.~Schuckers, R.~Singh, M.~Vatsa, A.~Noore, et~al.
\newblock Livdet iris 2017—iris liveness detection competition 2017.
\newblock In {\em 2017 IEEE International Joint Conference on Biometrics
  (IJCB)}, pages 733--741. IEEE, 2017.

\bibitem{zhao2017towards}
Z.~Zhao and A.~Kumar.
\newblock Towards more accurate iris recognition using deeply learned spatially
  corresponding features.
\newblock In {\em Proceedings of the IEEE International Conference on Computer
  Vision}, pages 3809--3818, 2017.

\end{thebibliography}
}

\end{document}